\documentclass[10pt,twocolumn,letterpaper]{article}
\usepackage{graphicx}

\usepackage{cvpr}
\usepackage{times}
\usepackage{epsfig}
\usepackage{graphicx}
\usepackage{amsmath}
\usepackage{amssymb}

\usepackage{makecell}
\usepackage{multirow}
\usepackage{bm}
\usepackage{placeins}
\usepackage{booktabs}

\usepackage{blindtext}
\usepackage{caption}
\newcommand{\mycomment}[1]{}

\usepackage[pagebackref=true,breaklinks=true,letterpaper=true,colorlinks,bookmarks=false]{hyperref}

\cvprfinalcopy 

\def\cvprPaperID{12} 

\ifcvprfinal\fi
\begin{document}

\newcommand{\refsec}[1]{Sec.~\ref{#1}}
\newcommand{\refsecb}[1]{Sec.~\ref{#1}~}

\newcommand{\reffig}[1]{Fig.~\ref{#1}}
\newcommand{\reftab}[1]{Table~\ref{#1}}
\newcommand{\ncomment}[1]{{\color{red}{[NN: #1]}}}
\newcommand{\icomment}[1]{{\color{blue}{[IK: #1]}}}
\newcommand{\acomment}[1]{{\color{green}{[ALP: #1]}}}
\title{DensePose: Dense Human Pose Estimation In The Wild}

\author{R{\i}za Alp G\"uler\footnotemark\\
INRIA-CentraleSup\'elec\\
{\tt\small riza.guler@inria.fr}
\and
Natalia Neverova\\
Facebook AI Research\\
{\tt\small nneverova@fb.com}
\and
Iasonas Kokkinos\\
Facebook AI Research\\
{\tt\small iasonask@fb.com} 
}

\twocolumn[{%
\maketitle
\vspace*{-10mm}
\begin{center}
    \centering
    \includegraphics[width=1\textwidth,trim={0 0 0 0},clip]{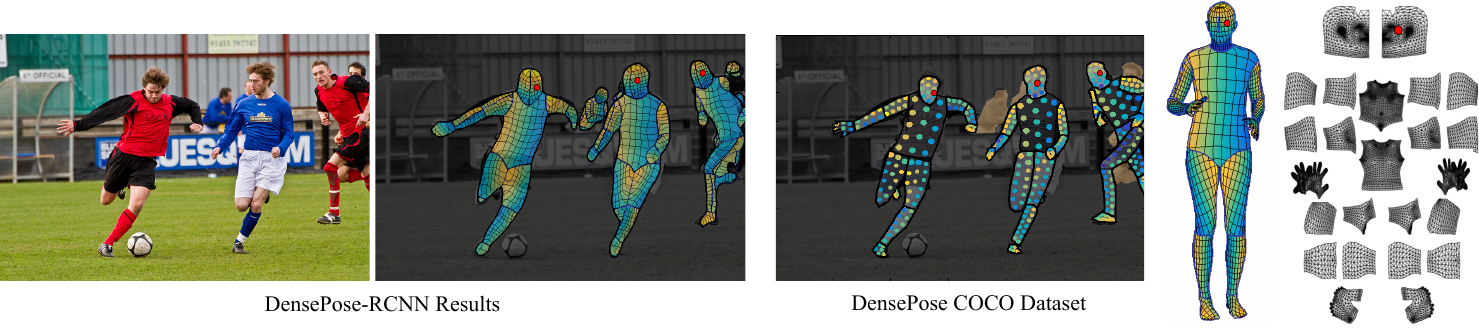}
    \captionof{figure}{
    Dense  pose estimation aims at mapping all human pixels of an RGB image to the 3D surface of the human body.
    We introduce DensePose-COCO, a large-scale  ground-truth dataset with image-to-surface  correspondences manually annotated on 50K COCO images and
	train DensePose-RCNN, to densely regress part-specific UV coordinates within every human region at multiple frames per second.  \textit{Left:} The image and the regressed correspondence by DensePose-RCNN, \textit{Middle:} DensePose COCO Dataset annotations, \textit{Right:} Partitioning and  UV parametrization of the  body surface.
    } 
    \label{fig:teaser}
\end{center}%
}]
\footnotetext[1]{R{\i}za Alp G\"uler was with Facebook AI Research during this work.}
\begin{abstract}

In this work, we establish dense correspondences between an RGB image and a surface-based representation of the human body, a task we refer to as dense human pose estimation.
We first gather dense  correspondences for 50K persons appearing in the  COCO dataset by introducing  an efficient annotation pipeline.
We then use our dataset to train CNN-based systems that  deliver dense correspondence `in the wild', namely in the presence of background, occlusions and scale variations.
We improve our training set's effectiveness by training an `inpainting' network that can fill in missing ground truth values, and report  clear improvements with respect to the best results that would be achievable in the past.
We experiment with fully-convolutional networks and region-based models and observe a superiority of the latter; we further improve accuracy through cascading, obtaining a system that delivers highly-accurate results in real time. Supplementary materials and videos are provided on the project page \url{http://densepose.org}.
\end{abstract}

\begin{figure*}[!h]
\centering
\includegraphics[ width=0.98\linewidth ]{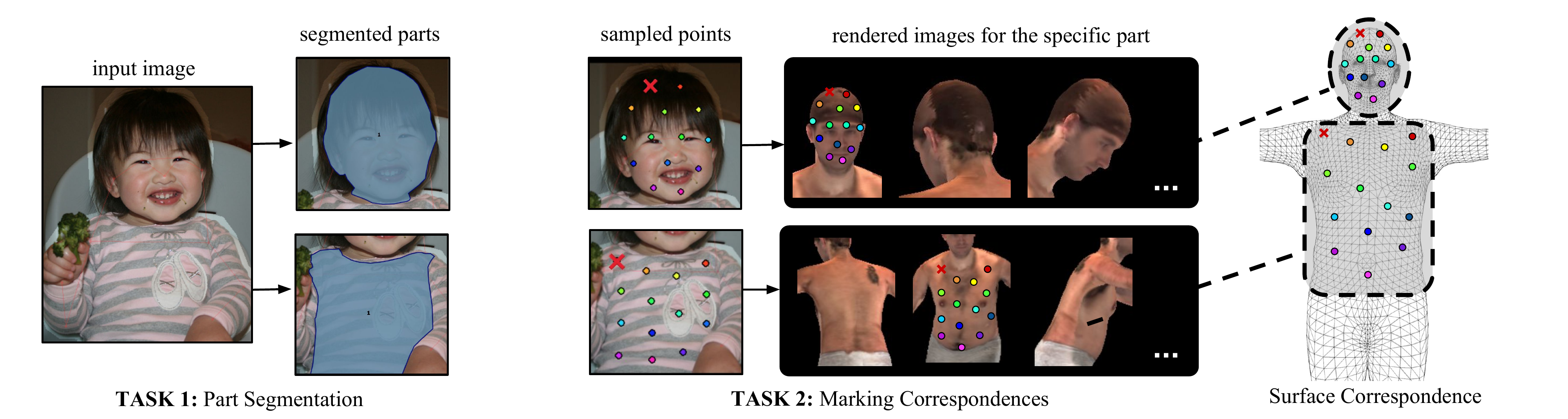}
\caption{We  annotate dense correspondence between images and a 3D surface model by asking the annotators to segment the image into semantic regions and  to then
localize the corresponding surface point for each of the sampled points on any of the rendered part images. The red cross indicates the currently annotated point.
The surface coordinates of the rendered views  localize the collected 2D points on the  3D model.
}
\label{fig:Pipeline}
\end{figure*}

\section{Introduction}

This work aims at pushing further the envelope of human understanding in images by establishing dense correspondences from a 2D image to a 3D, surface-based representation of the human body. We can understand this task as involving several other problems, such as object detection, pose estimation, part and instance segmentation  either as  special cases or prerequisites. Addressing this task has applications in problems that require going beyond plain landmark localization, such as graphics, augmented reality, or human-computer interaction, and could also be a stepping stone towards general 3D-based object understanding.

The task of establishing dense correspondences from an image to a surface-based  model has been addressed  mostly in the setting where a depth sensor is available, as in the Vitruvian manifold  of \cite{TaylorSSF12}, metric regression forests \cite{pons2015metric}, or the more recent dense point cloud correspondence  of \cite{WeiHCVL16}. By contrast, in our case we consider a single RGB image as input, based on which we establish a correspondence between surface points and image pixels.

Several other works have recently aimed at recovering dense correspondences between pairs \cite{BristowVL15} or sets of RGB images  \cite{ZhouKAHE16,denseiccv17} in an unsupervised setting.
More recently,~\cite{ThewlisBV17a} used the equivariance principle in order to align sets of images to a common coordinate system, while following the general idea of groupwise image alignment, e.g.~\cite{congealing,iccv07}. 

While these works are aiming at general categories, our work is focused on arguably the most important visual category, humans. 
For humans one can simplify the task by exploiting  parametric deformable surface models, such as the Skinned Multi-Person Linear (SMPL) model of  ~\cite{smpl3d}, or the more recent Adam model of \cite{Sheikh18} obtained through carefully controlled 3D surface acquisition. Turning to the task of image-to-surface mapping, in ~\cite{smpl3d}, the authors propose a two-stage method of first detecting human landmarks through a CNN and then fitting a parametric deformable surface model to the image through iterative minimization. In parallel to our work, \cite{hmrKanazawa17} develop the method of \cite{smpl3d} to operate in an end-to-end fashion, incorporating the iterative reprojection error minimization as a module of a deep network that recovers 3D camera pose and the low-dimensional body parametrization. 

Our methodology differs from all these  works in that we take a full-blown supervised learning approach and gather ground-truth correspondences between images and a detailed, accurate parametric surface model of the human body \cite{smpl}: rather than using the SMPL model at test time we only use it as a means of defining our problem during training. 
Our approach can be understood as the next step in the line of works on 
extending the standard for humans in \cite{coco,andriluka14cvpr,Johnson10,dantone2014body,sigal2010humaneva,h36m_pami,mocap2003data}.
Human part segmentation masks have been provided in the
Fashionista~\cite{yamaguchi2012parsing}, PASCAL-Parts~\cite{chen2014detect}, 
and  Look-Into-People (LIP)~\cite{gong2017look} datasets; these can be understood as providing a coarsened version of image-to-surface correspondence, where rather than continuous coordinates one predicts  discretized part labels. 
Surface-level supervision was only recently introduced for synthetic images in \cite{varol2017learning}, while in \cite{UP} a dataset of 8515 images is annotated with keypoints and semi-automated fits of 3D models to images. In this work instead of compromising the extent and realism of our training set we introduce a novel annotation pipeline that allows us to gather ground-truth correspondences for  50K images of the COCO dataset, yielding our new DensePose-COCO dataset.

Our work is closest in spirit  to the recent DenseReg framework~\cite{densereg}, where CNNs were trained to successfully establish dense correspondences between a 3D model and images `in the wild'. That work focused mainly on  faces, and evaluated their results on datasets with moderate pose variability.
Here, however, we are facing new challenges, due to the higher complexity and flexibility of the human body, as well as the larger variation in poses. We address these challenges by designing appropriate architectures, as described in \refsec{method}, which yield substantial improvements over a DenseReg-type fully convolutional architecture. By combining our approach with the recent Mask-RCNN system of \cite{maskRCNN}
we show that a discriminatively trained model can recover highly-accurate correspondence fields for complex scenes involving tens of persons with real-time speed: on a GTX 1080 GPU our system operates at 20-26 frames per second for a $240 \times 320$ image or 4-5 frames per second for a $800 \times 1100$ image. 

Our contributions can be summarized in three points.
Firstly, as described in~\refsec{supervision}, 
we introduce the first manually-collected ground truth dataset for the task, by gathering dense  correspondences between the SMPL model~\cite{smpl} and persons appearing in the  COCO dataset. This is accomplished through  a novel annotation pipeline that exploits 3D surface information during annotation.

Secondly, as described in~\refsec{method}, we use the resulting dataset to train CNN-based systems that  deliver dense correspondence `in the wild', by regressing body surface coordinates at any image pixel. We experiment with both fully-convolutional architectures, relying on  Deeplab~\cite{deeplab}, and also with region-based systems, relying on  Mask-RCNN~\cite{maskRCNN},  observing  a  superiority of region-based models over fully-convolutional networks.
We also consider cascading variants of our approach, yielding further improvements over existing architectures.

Thirdly,  we explore different ways of exploiting our constructed ground truth information. Our supervision signal is defined over a randomly chosen subset of image pixels per training sample. We use these sparse correspondences to train   a `teacher' network that can `inpaint' the supervision signal in the rest of the image domain. Using this inpainted signal results in clearly better performance when compared to either sparse points, or any other existing dataset, as shown experimentally in~\refsec{experiments}.

Our experiments indicate that dense human pose estimation is to a large extent feasible, but still has space for improvement. We conclude our paper with some qualitative results and directions that show the potential of the method. We will make code and data publicly available from our project's webpage, \url{http://densepose.org}. 

\section{COCO-DensePose Dataset}
\label{supervision}

\mycomment{
\begin{figure*}[t]
\centering
\includegraphics[ width=0.95\linewidth ]{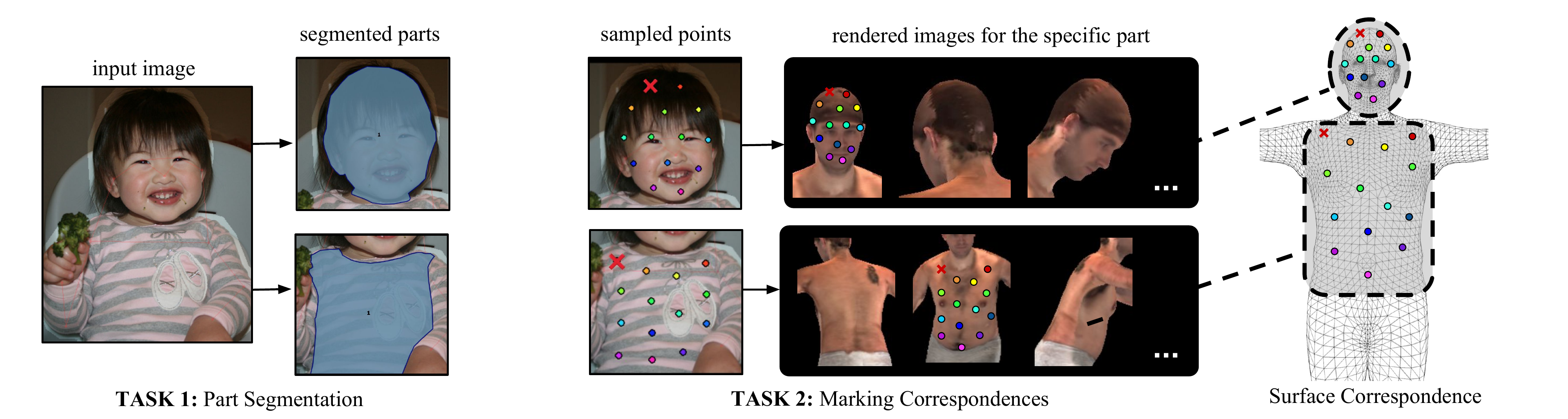}
\caption{ Our two-stage method for collecting annotations for correspondence between pixels and a 3d surface model: The first annotation task consists in the segmentation of semantic human parts. The second annotation task consists in localizing points sampled from each part  on one of six synthetic images rendered for that specific part (we show only three).  The  points collected on the rendered images are directly localized on the 3D model.}
\vspace{-0.35cm}
\label{fig:AnnotPipeline}
\end{figure*}
}

Gathering rich, high-quality training sets has been a catalyst for progress in the classification \cite{imagenet}, detection and segmentation \cite{PASCAL,coco} tasks. There currently exists no manually collected ground-truth  for dense human pose estimation for real images. The works of \cite{UP} and \cite{varol2017learning} can be used as surrogates, but as we show in \refsecb{experiments}  provide worse supervision. 

\begin{figure}[!t]
\centering
\includegraphics[ width=1\linewidth ]{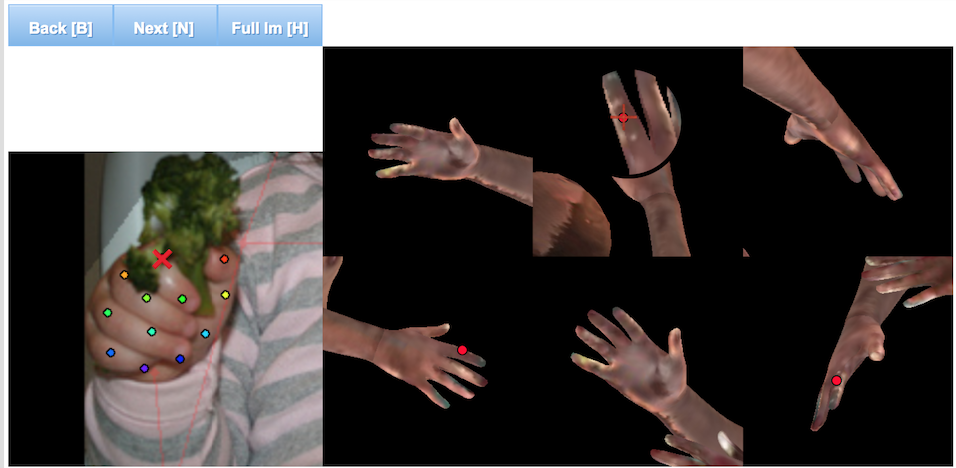}
\caption{ The user interface for collecting per-part correspondence annotations: We provide the annotators six pre-rendered views of a body part such that the whole part-surface is visible. Once the target point is annotated, the point is displayed on all rendered images simultaneously. }
\label{fig:correspondence_interface}
\end{figure}

In this Section we introduce  our COCO-DensePose dataset, alongside with evaluation measures that allow us to quantify progress in the task in \refsec{experiments}. We have gathered annotations for 50K humans, collecting more then 5 million manually annotated correspondences. 

We start with a  presentation of our annotation pipeline, since this required several design choices that may be more generally useful for 3D annotation. We then turn to an analysis of the accuracy of the gathered ground-truth, alongside with the resulting performance measures used to assess the different methods.

\begin{figure*}[!t]
\centering
\includegraphics[width=1.\textwidth]{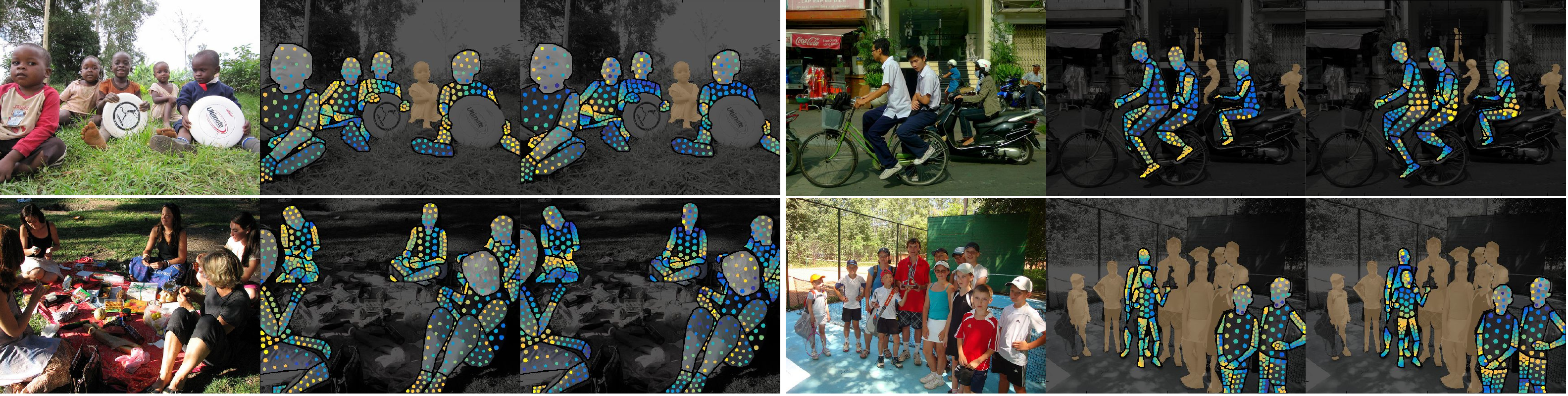}
\caption{ Visualization of annotations: Image (left), U (middle) and V (right) values for the collected points.}
\label{fig:anno1}
\end{figure*}

\subsection{Annotation System}
In this work, we involve human annotators to establish dense correspondences from 2D images to surface-based representations of the human body. If done naively, this would require `hunting vertices' for every 2D image point, by manipulating a surface through rotations - which can be frustratingly inefficient.
Instead, we construct an annotation pipeline through which we can efficiently gather annotations for image-to-surface correspondence. 


As shown in~\reffig{fig:Pipeline}, in the first stage we ask  annotators to delineate regions corresponding to visible, semantically defined  body parts.
These include Head, Torso, Lower/Upper Arms, Lower/Upper Legs, Hands and Feet. 
In order to use simplify the UV parametrization 
we design the parts to be isomorphic to a plane,  partitioning the limbs and torso into lower-upper and frontal-back parts. 

For \textit{head}, \textit{hands} and \textit{feet}, we use the manually obtained UV fields provided in the SMPL model~\cite{smpl}. For the rest of the parts we obtain the unwrapping via multi-dimensional scaling applied to pairwise geodesic distances. The UV fields for the resulting 24 parts are visualized in Fig.~\ref{fig:teaser}~(right).

We  instruct the annotators to estimate the body part behind the clothes, so that for instance wearing a large skirt  would not complicate the subsequent annotation of correspondences. In the second stage we  sample every part region with a set of roughly equidistant points obtained via k-means and request the annotators to bring these points in correspondence with the surface. The number of sampled points varies based on the size of the part and the maximum number of sampled points per part is 14. In order to simplify this task  we `unfold' the part surface by providing six pre-rendered views of the same body part and allow the user to place landmarks on any of them \reffig{fig:correspondence_interface}. This allows the annotator to choose  the most convenient point of view by selecting one among six options instead of manually rotating the surface. 

As the user indicates a point on any of the rendered part views, its surface coordinates are used to simultaneously show its position on the remaining views -- this gives a global overview of the correspondence. The image points are presented to the annotator in a horizontal/vertical succession, which makes it easier to deliver geometrically consistent annotations by avoiding self-crossings of the surface.
This two-stage annotation process has allowed us to very efficiently gather highly accurate correspondences. If we quantify the complexity of the annotation task in terms of the time it takes to complete it, we have seen that the part segmentation and correspondence annotation tasks take approximately the same time, which is  surprising given the more challenging nature of the latter task. Visualizations of the collected annotations are provided in Fig.~\ref{fig:anno1}, where the partitioning of the surface and U, V coordinates are shown in Fig.~\ref{fig:teaser}.

\newcommand{\refeq}[1]{Eq.~{\ref{#1}}}

\subsection{Accuracy of human annotators}

We assess human annotator with respect to a gold-standard measure of performance. 
Typically in pose estimation one asks multiple annotators to label the same landmark, which is then used to assess the variance in position, e.g. \cite{coco,Ronchi_2017_ICCV}. In our case, we can render images where we have access to the true mesh coordinates used to render a pixel. We thereby directly compare the true position used during rendering and the one estimated by annotators, rather than first estimating a 'consensus' landmark location  among multiple human annotators.

\newcommand{\gtidx}[1]{{#1}}
\newcommand{\gtest}[1]{\hat{#1}}
\newcommand{\geod}[2]{d_{#1,#2}}
\newcommand{\geoderr}[2]{d_{#1,#2}}
\newcommand{\geodest}[1]{\bar{d}_{#1}}
\newcommand{\geodinterp}[2]{\hat{d}_{#1,#2}}
\newcommand{\sset}[1]{\mathcal{S}_{#1}}
\newcommand{\simil}[1]{k(#1)}
\newcommand{\beq}{\begin{equation}}
\newcommand{\eeq}{\end{equation}}

In particular, we  provide annotators with synthetic images generated through the exact same surface model as the one we use in our ground-truth annotation, exploiting the rendering system and textures of~\cite{varol2017learning}. 
We then ask annotators to bring the synthesized images into correspondence with the surface using our annotation tool, and for every image~$k$ estimate the  geodesic distance~$\geoderr{i}{k}$ between the correct surface point, $\gtidx{i}$ and the point estimated by human annotators~$\gtest{i}_k$:
\beq
\geoderr{i}{k} = g(\gtidx{i},\gtest{i}_k),
\eeq
where $g(\cdot,\cdot)$ measures the geodesic distance between two surface points.

For any image~$k$, 
we annotate and estimate the error only on a randomly sampled set of surface points~$\sset{k}$ and interpolate the errors on the remainder of the surface.
Finally, we average the errors across all $K$ examples used to assess annotator performance.

\begin{figure}[!b]
\centering
\includegraphics[ width=1\linewidth ]{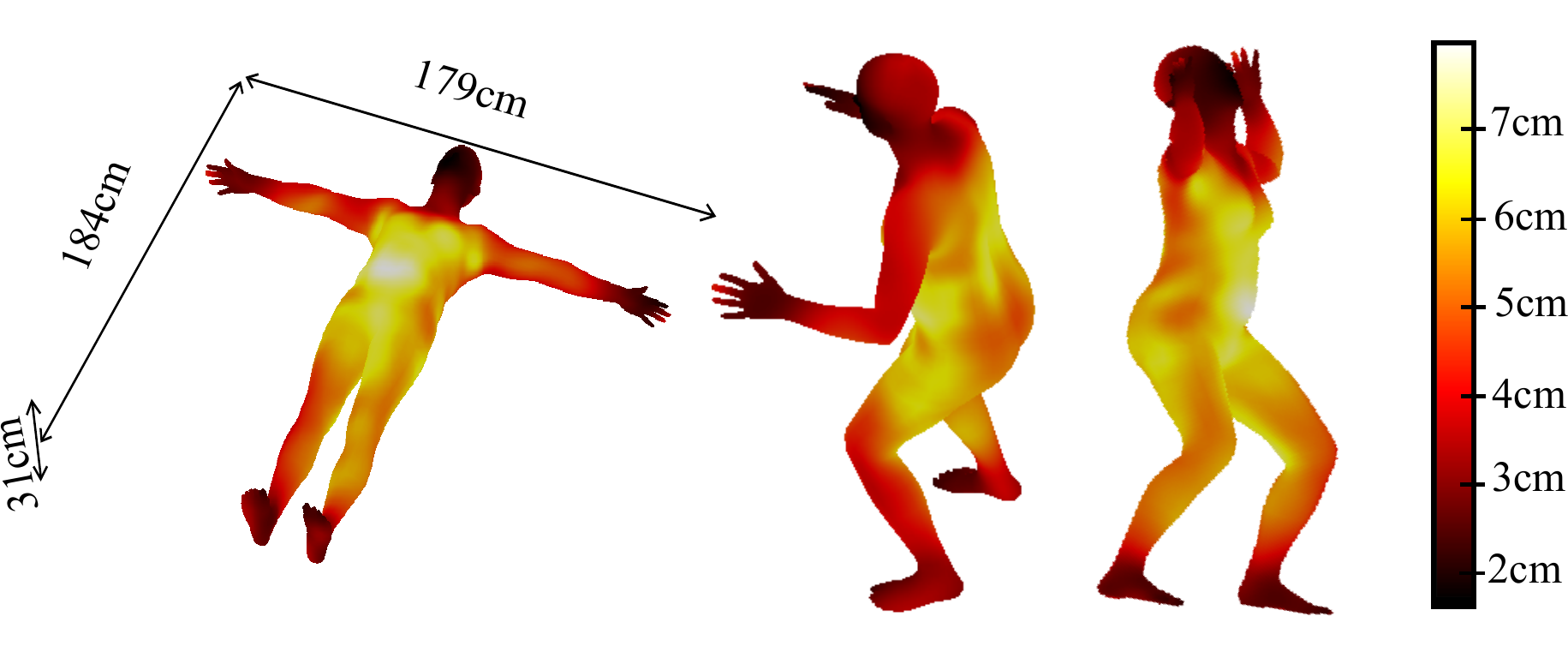}
\caption{ Average human annotation error as a function of surface position.}
\label{fig:errors}
\end{figure}

As shown in \reffig{fig:errors}~ the annotation errors are substantially smaller on small surface parts with distinctive features that could help localization (face, hands, feet), while on larger uniform areas that are typically covered by clothes (torso, back, hips) the annotator errors can get larger. 

\subsection{Evaluation Measures}
\label{measures}

\begin{figure}[!b]
\centering
\includegraphics[ width=0.93\linewidth ]{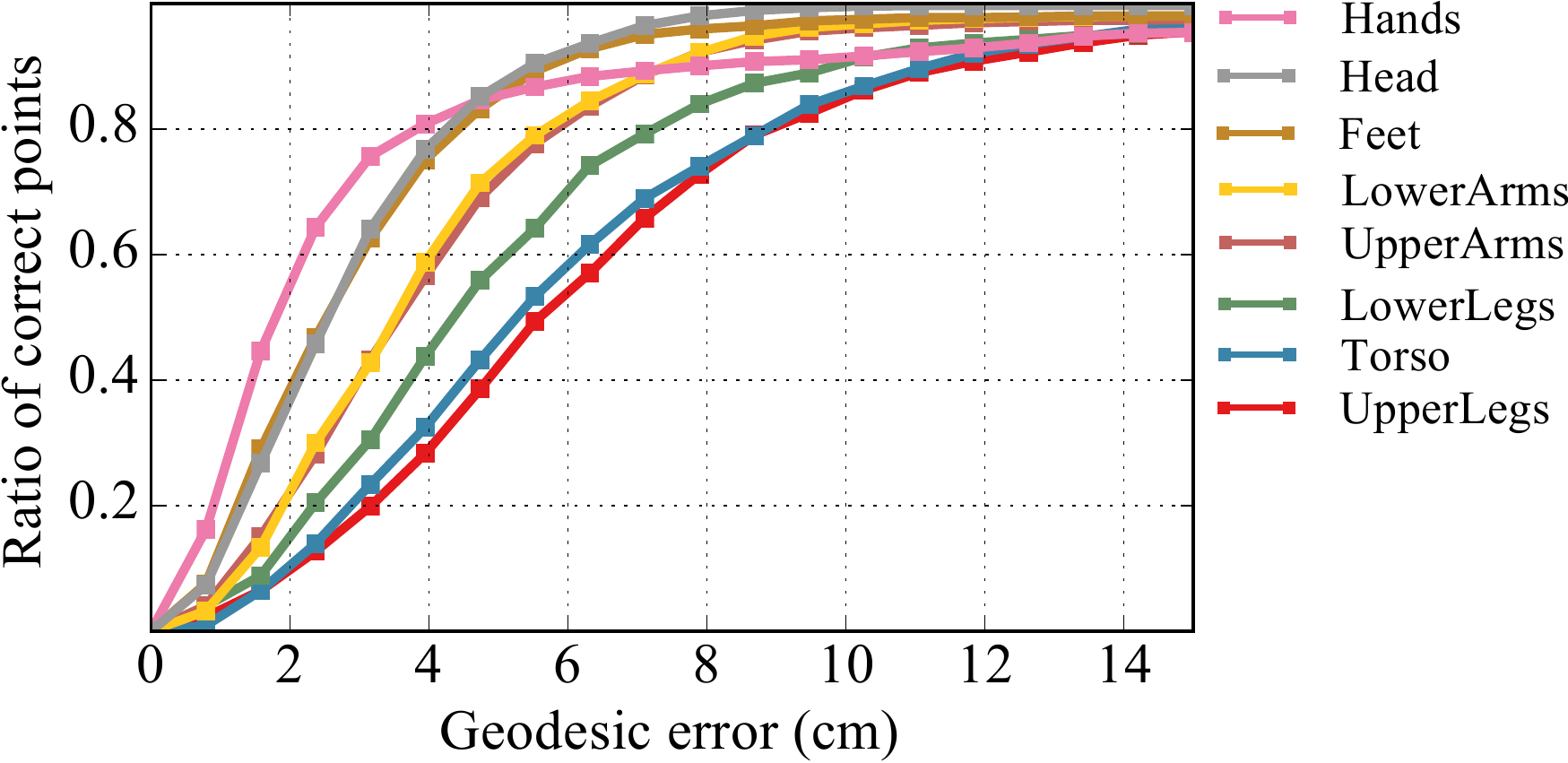}
\caption{ Human annotation error distribution within different body parts.}
\label{fig:perrors}
\end{figure}

We consider two different ways of summarizing correspondence accuracy over the whole human body, including pointwise and per-instance evaluation. 
\paragraph{Pointwise evaluation.} This approach evaluates correspondence accuracy over the whole image domain through the Ratio of Correct Point (RCP) correspondences, 
where a correspondence is declared correct if the geodesic distance is below a certain threshold. As the threshold $t$ varies, we obtain a curve $f(t)$, whose area provides us with a scalar summary of the correspondence accuracy. 
For any given image we have a varying set of points coming with ground-truth signals. We summarize performance on the ensemble of such points, gathered across images.
We evaluate the area under the curve (AUC),  $\mathrm{AUC}_{a} = \frac{1}{a}\int_0^a f(t) dt$, for two different values of $a=10 \mathrm{cm}, 30 \mathrm{cm}$ yielding $\text{AUC}_{10}$ and $\text{AUC}_{30}$ respectively, where $\text{AUC}_{10}$ is understood as being an accuracy measure for more refined correspondence.
This performance measure 
is easily applicable to both single- and multi-person scenarios and can deliver directly comparable values. In \reffig{fig:perrors}, we provide the per-part pointwise evaluation of the human annotator performance on synthetic data, which can be seen as an upper bound for the performance of our systems.

\paragraph{Per-instance evaluation.} Inspired by the object keypoint similarity~(OKS) measure used for pose evaluation on the COCO dataset \cite{coco, Ronchi_2017_ICCV}, 
 we introduce \textit{geodesic point similarity~(GPS)}\ as a correspondence matching score:
\begin{equation}
\text{GPS}_j = \frac{1}{|P_j|}\sum_{p \in P_j}\exp\left(\frac{-g(\gtidx{i_p},\gtest{i}_p)^2}{2\kappa^2}\right),
\end{equation}
where $P_j$ is the set of ground truth points 
annotated on person instance~$j$, $\gtidx{i_p}$ is the vertex estimated by a model at point~$p$, $\gtest{i}_p$ is the ground truth vertex~$p$ and~$\kappa$ is a normalizing parameter.
We set  $\kappa{=}0.255$  so  that a single point has a $\text{GPS}$ value of $0.5$ if its geodesic distance from the ground truth equals the average half-size of a body segment, corresponding to approximately $30$\,cm. 
Intuitively, this means that a score of $\text{GPS}\,{\approx}\,0.5$ can be achieved by a perfect part segmentation model, while going above that also requires  a more precise localization of a point on the surface. 

Once the matching is performed, we follow the COCO challenge protocol~\cite{coco,poseeval} and evaluate Average Precision (AP) and Average Recall (AR) at a number of GPS thresholds ranging from 0.5 to 0.95, which corresponds to the range of geodesic distances between $0$ and $30$\,cm. We use the same range of distances to perform both per-instance and per-point evaluation.

\mycomment{
Note: here the geodesic distance is set to inf if 1) the point is classified as background, 2) the point is outside of the detected box.
Maybe here we can put figure with geodesic distances and here we can draw our groundtruth as annotated by our annotators. Also we can put here some dummy baseline, obtained by pose -  selecting random vertices, etc..
}

\section{Learning Dense Human Pose Estimation}
\label{method}
We now turn to the task of training a deep network that predicts dense correspondences between image pixels and surface points.
Such a task was recently addressed in the Dense Regression (DenseReg) system of~\cite{densereg} through a fully-convolutional network  architecture~\cite{deeplab}. In this work, we introduce improved architectures  by combining the DenseReg approach with the Mask-RCNN architecture~\cite{maskRCNN}, yielding our `DensePose-RCNN' system. We  develop cascaded extensions of DensePose-RCNN that further improve accuracy and describe a training-based interpolation method that allows us to turn a sparse supervision signal into a denser and more effective variant.

\subsection{Fully-convolutional dense pose regression}
The simplest architecture choice consists in using a fully convolutional network (FCN) that combines a classification and a regression task, similar to DenseReg.
In a first step, we classify a pixel as belonging to either background, or one among several region parts which provide a coarse estimate of surface coordinates. This amounts to a labelling task that is trained using a standard cross-entropy loss.
In a second step, a regression system indicates the exact coordinates of the pixel within the part. 
Since the human body has a complicated structure, we  break it into multiple independent pieces and parameterize each piece using a local two-dimensional coordinate system, that identifies the position of any node on this surface part.

\newcommand{\argmax}{\mathrm{argmax}}
Intuitively, we can say that we first use appearance to make a coarse estimate of where the pixel belongs to and then align it to the exact position through some small-scale correction.
Concretely, coordinate regression at an image position $i$ can be formulated as follows:
\begin{gather}
c^{\ast} = \argmax_{c}P(c|i), \quad
[U,V] = R^{c^{\ast}}(i)
\end{gather}
where in the first stage we assign position~$i$ to the body part~$c^{\ast}$ that has highest posterior probability, as calculated by the classification branch, and in the second stage we use the regressor $R^{c^{\ast}}$ that  places the point~$i$ in the continuous $U,V$  coordinates parametrization of part $c^{\ast}$. In our case, $c$ can take 25 values (one is background), meaning that $P_x$ is a 25-way classification unit, and we train 24 regression functions $R^{c}$, each of which provides 2D coordinates within its respective part $c$. While training, we use a cross-entropy loss for the part classification and a smooth $L_1$ loss for training each regressor. The regression loss is only taken into account for a part if the pixel is within the specific part.

\subsection{Region-based Dense Pose Regression} 

\begin{figure}[t]
\centering
\includegraphics[ width=0.95\linewidth ]{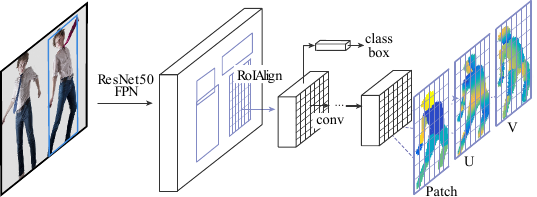}
\caption{ DensePose-RCNN architecture: we use a cascade of region proposal generation and feature pooling,  followed by a fully-convolutional network that densely predicts discrete part labels and continuous surface coordinates.}

\vspace{-0.05cm}
\label{fig:MaskRcnn}
\end{figure}

Using an FCN makes the system particularly easy to train, but loads the same deep network with too many tasks, including part segmentation and  pixel localization, while at the same time requiring scale-invariance which becomes challenging for humans in COCO. 
Here we adopt the region-based approach of \cite{fasterRCNN,maskRCNN}, which consists in a cascade of proposing  regions-of-interest (ROI), extracting region-adapted features through ROI pooling \cite{spp,maskRCNN} and feeding the resulting features into a region-specific branch.
Such architectures decompose the complexity of the task into controllable modules and  implement a scale-selection mechanism through ROI-pooling.
At the same time, they can also be trained jointly in an end-to-end manner \cite{fasterRCNN}. 

We adopt the settings introduced in~\cite{maskRCNN}, involving the construction of Feature Pyramid Network \cite{fpn} features, and ROI-Align pooling, which have been shown to be important for tasks that require spatial accuracy. We adapt this architecture to our task, so as to obtain dense part labels and coordinates within each of the selected regions. 

As shown in \reffig{fig:MaskRcnn}, we introduce a fully-convolutional network on top of ROI-pooling  that is entirely devoted to these two tasks, generating a classification and a regression head that provide the part assignment and part coordinate predictions, as in DenseReg. For simplicity, we use the exact same architecture used in the keypoint branch of Mask-RCNN, consisting of a stack of 8 alternating  $3{\times}3$ fully convolutional and ReLU layers with 512 channels.
At the top of this branch we have the same classification and regression losses as in the FCN baseline, but we now use a supervision signal that is cropped within the proposed region. 

During inference, our system operates at 25fps on 320x240 images and 4-5fps on 800x1100 images using a GTX1080 graphics card.

\subsection{Multi-task cascaded architectures}
\label{cascade}

Inspired by the success of recent pose estimation models based on iterative refinement~\cite{CPM,hourglass} we experiment with cascaded architectures. Cascading can improve performance both by providing context to the following stages, and also through the benefits of deep supervision \cite{dsn}.

\begin{figure}[!t]
\centering
\includegraphics[ width=\linewidth ]{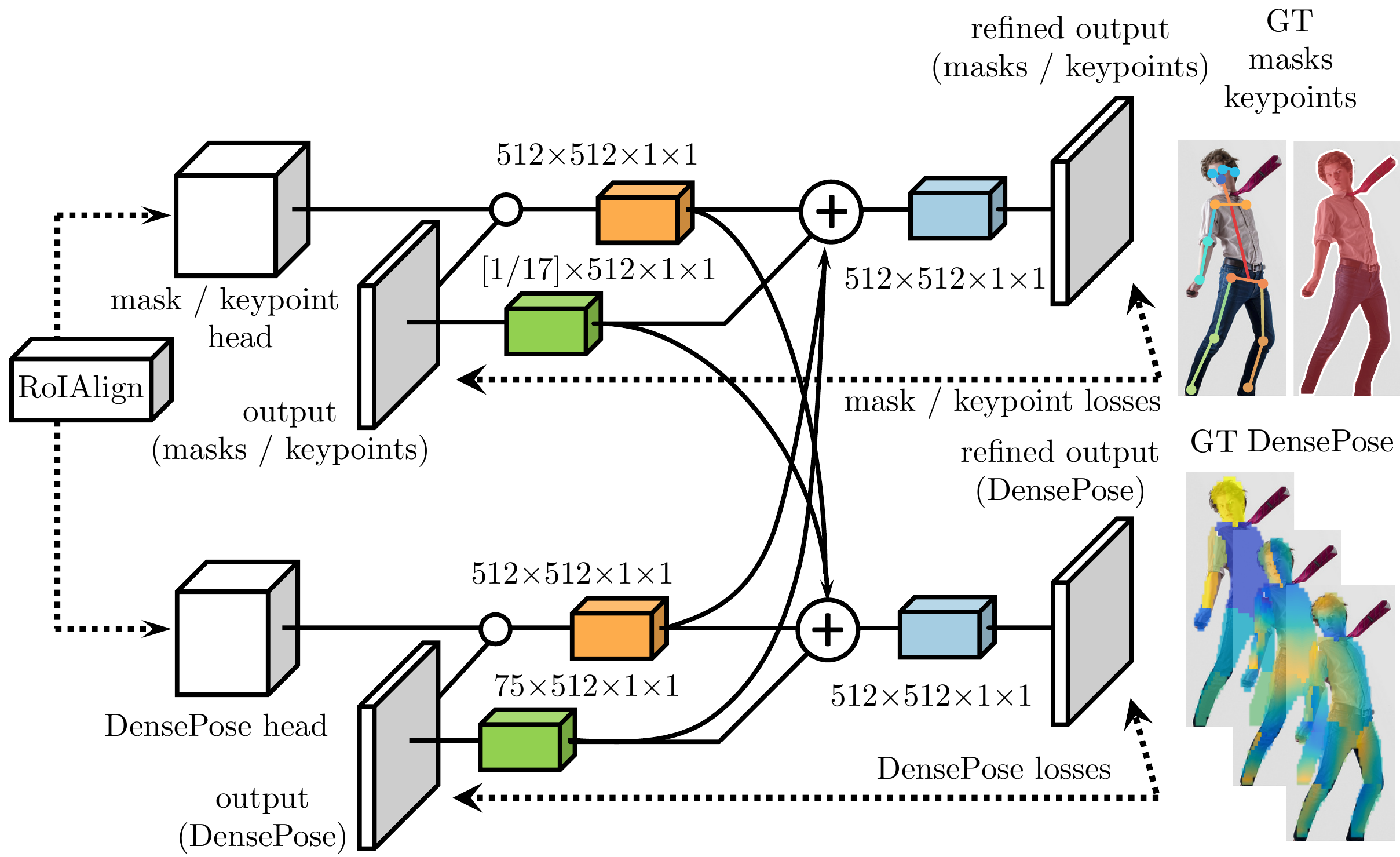}
\caption{ Cross-cascading architecture: The output of the RoIAlign module in \reffig{fig:MaskRcnn} feeds into the DensePose network as well as auxiliary networks for other tasks (masks, keypoints). Once first-stage predictions are obtained from all tasks, they are combined and then fed into a second-stage refinement unit of each branch.}
\vspace{-0.05cm}
\label{fig:Cascade}
\end{figure}

As shown in \reffig{fig:Cascade}, we do not confine ourselves to cascading within a single task, but also exploit  information from related tasks, such as keypoint estimation and instance segmentation, which have successfully been addressed by the Mask-RCNN architecture \cite{maskRCNN}. This allows us to exploit task synergies and the complementary merits of different sources of supervision.

\subsection{Distillation-based ground-truth interpolation} 
\label{distillation}
\begin{figure}[!b]
\centering
\includegraphics[ width=0.95\linewidth ]{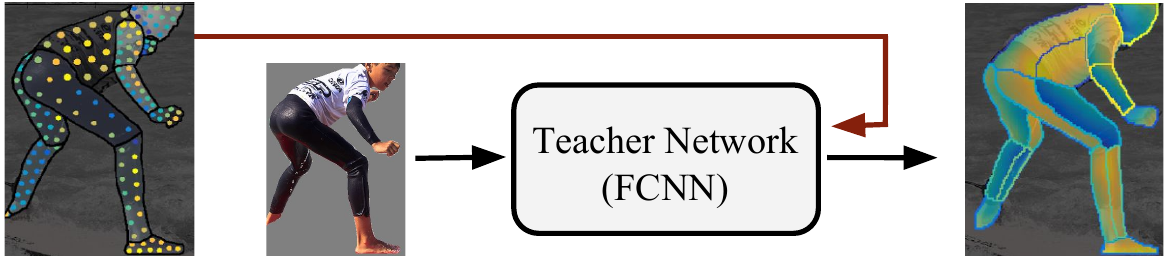}
\caption{ We first train a `teacher network' with our sparse, manually-collected supervision signal, and then use the network to `inpaint' a dense supervision signal used to train our region-based system. }
\vspace{-0.05cm}
\label{fig:teacher}
\end{figure}

Even though we aim at dense pose estimation at test time, in every training sample we annotate only a sparse subset of the pixels,
approximately 100-150 per human.
This does not necessarily pose a problem during training, since we can make our classification/regression losses oblivious to points where the ground-truth correspondence was not collected, simply by not including them in the summation over the per-pixel losses \cite{fcn}.
However, we have observed that we obtain substantially better results by ``inpainting'' the values of the supervision signal on positions that were not originally annotated. For this  we adopt a learning-based approach where we firstly train a ``teacher'' network (depicted in \reffig{fig:teacher}) to reconstruct the ground-truth values wherever these are observed, and then deploy it on the full image domain, yielding a dense supervision signal. In particular, we only keep the network's predictions on areas that are labelled as foreground, as indicated by the part masks collected by humans, in order to ignore network errors on background regions. 

\begin{figure*}[!t]
\centering

\includegraphics[ width=1\linewidth]{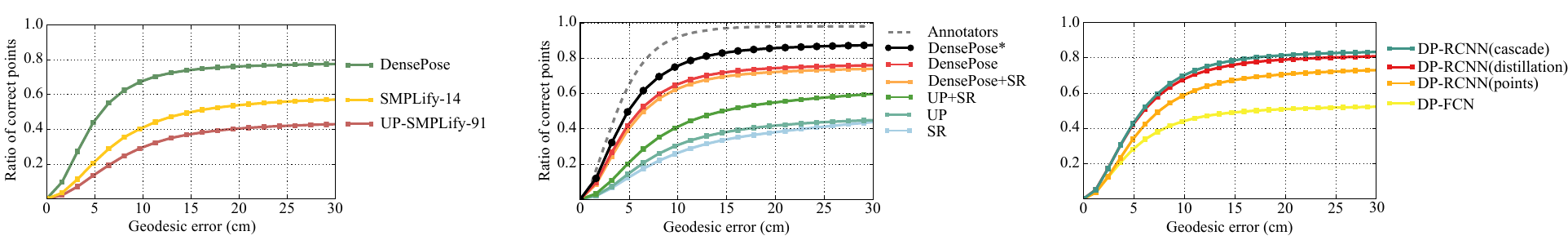}
\vspace{-0.6cm}
\label{fig:Point_errors}
\end{figure*}

%
\begin{figure*}[!htb]
\minipage{0.29\textwidth}
\centering
\scalebox{0.72}{
\begin{tabular}{|l|c|c|c|}
\hline
$\vphantom{\rule{0mm}{3.5mm}}$
\hspace*{8.8mm}\emph{Method} & \hspace*{0.1mm}\textbf{\emph{$\mathbf{AUC}_{10}$} }\hspace*{0.1mm}&  \hspace*{0.1mm}\emph{$\mathbf{AUC}_{30}$}\hspace*{0.1mm} \\
\hline\hline
\multicolumn{3}{|c|}{$\vphantom{\rule{0mm}{3.5mm}}$ 
\textit{Full-body images}}\\ \hline
$\vphantom{\rule{0mm}{3.5mm}}$%
UP-SMPLify-91 &  0.155 & 0.306\\
SMPLify-14 &  0.226 & 0.416\\
DensePose &  0.429 & 0.630\\
\hline\hline
\multicolumn{3}{|c|}{$\vphantom{\rule{0mm}{3.5mm}}$ 
\textit{All images}}\\ \hline
$\vphantom{\rule{0mm}{3.5mm}}$%
SMPLify-14      &  0.099 &0.19 \\
DensePose      &  0.378 &0.614\\
\hline\hline
$\vphantom{\rule{0mm}{3.5mm}}$%
Human Performance\hspace*{2.3mm} &  0.563 &0.835 \\
\hline
\end{tabular}
}\vspace*{-0.9mm}
\caption{ Qualitative comparison between model-based single-person pose estimation of SMPLify~\cite{smpl3d} and our FCN-based result, in the absence (`full-body images') and presence (`all images') of occlusions.} 

\label{fig:SMPLFY}
\endminipage\hfill
\minipage{0.31\textwidth}
\centering
\scalebox{0.72}{
\begin{tabular}{|l|c|c|c|}
\hline
$\vphantom{\rule{0mm}{3.5mm}}$
\hspace*{8.8mm}\emph{Method} & \hspace*{1.01mm}\textbf{\emph{$\mathbf{AUC}_{10}$} }\hspace*{1.01mm}\,&  \hspace*{1.01mm}\emph{$\mathbf{AUC}_{30}$}\hspace*{1.01mm}\, \\
\hline\hline
$\vphantom{\rule{0mm}{3.5mm}}$%
SR             &  0.124 &0.289\\
UP             &  0.146 &0.319\\
SR + UP        &  0.201 &0.424 \\
DensePose + SR &  0.357 &0.592 \\
DensePose      &  0.378 &0.614\\
DensePose$^*$  &  0.445 &0.711\\
\hline\hline
$\vphantom{\rule{0mm}{3.5mm}}$%
Human Performance\hspace*{2.5mm} &  0.563 &0.835 \\
\hline
\end{tabular}\vspace*{0.5mm}
}
\caption{
Single-person performance for different kinds of supervision signals used for training: DensePose leads to substantially more accurate results than surrogate datasets. DensePose$^{\ast}$  uses a figure-ground oracle at both  training and test time.}
\label{fig:datasets}
\endminipage\hfill
\minipage{0.34\textwidth}
\centering
\scalebox{0.7}{
\begin{tabular}{|l|c|c|c|}
\hline
$\vphantom{\rule{0mm}{3.5mm}}$
\hspace*{12mm}\emph{Method} & \textbf{\emph{$\mathbf{AUC}_{10}$} }&  \emph{$\mathbf{AUC}_{30}$} &\emph{$\mathbf{IoU}$}   \\
\hline\hline
$\vphantom{\rule{0mm}{3.5mm}}$%
DP-FCN          &  0.253 & 0.418 & 0.66\\
DP-RCNN (points only)  & 0.315    &  0.567 & 0.75\\
DP-RCNN (distillations)  &  0.381 &  0.645 &0.79 \\
DP-RCNN (cascade)  &  0.390 &  0.664 &0.81\\
DP$^*$  &  0.417 &0.683&$-$\\
\hline\hline
$\vphantom{\rule{0mm}{3.5mm}}$%
Human Performance &  0.563 &0.835 &$-$\\
\hline
\end{tabular}
}\vspace*{-2pt}
\caption{ Results of multi-person dense correspondence labelling. Here we compare the performance of our proposed DensePose-RCNN  system against the fully-convolutional alternative on realistic images from the COCO dataset including multiple persons with high variability in scales, poses and backgrounds.}
\label{fig:dense_errors}
\endminipage
\end{figure*}

\section{Experiments}
\label{experiments}
In all of the following experiments, we assess the methods on a test set of 1.5k images containing 2.3k humans, using as training set of 48K humans. Our test-set coincides with the COCO keypoints-minival partition used by~\cite{maskRCNN} and the training set with the COCO-train partition. We are currently collecting annotations for the remainder of the COCO dataset, which will soon allow us to also have a competition mode evaluation. 

Before assessing dense pose estimation `in the-wild' in \refsec{secondpart}, we start in \refsec{firstpart} with the more restricted `Single-Person' setting where we use as inputs images cropped around ground-truth boxes. This factors out the effects of detection performance and provides us with a controlled setting to assess the usefulness of the COCO-DensePose dataset.

\subsection{Single-Person Dense Pose Estimation}
\label{firstpart}

We start in \refsecb{super} by comparing the COCO-DensePose dataset to other sources of supervision for dense pose estimation 
and then in \refsecb{modelc} compare the performance of the model-based system of \cite{smpl3d} with our discriminatively-trained system. Clearly the system of \cite{smpl3d} was not trained with the same amount of data as our model; this comparison therefore serves primarily to show the merit of our large-scale dataset for discriminative training.

\subsubsection{Manual supervision versus surrogates}
\label{super}
We start by assessing whether COCO-DensePose improves the accuracy of dense pose estimation with respect to the prior semi-automated, or synthetic supervision signals described below. 

A semi-automated method is used for  the `Unite the People' (UP) dataset of~\cite{UP}, where human annotators  verified the results of fitting the SMPL 3D deformable model~\cite{smpl}
to 2D images.
However, model fitting often fails in the presence of occlusions, or extreme poses, and is never guaranteed to be entirely successful -- for instance, even after rejecting a large fraction of the fitting results, the feet are still often misaligned in~\cite{UP}. This both decimates the training set and obfuscates evaluation, since the ground-truth itself may have systematic errors. 

Synthetic ground-truth can be established by rendering images using surface-based models ~\cite{pishchulin2011learning,pishchulin2012articulated,rogez2016mocap,ghezelghieh2016learning,chen2016synthesizing,neverova2017hands}. This has recently been applied to human pose in the SURREAL dataset of \cite{varol2017learning}, where  the SMPL model ~\cite{smpl} was rendered with the CMU Mocap dataset poses~\cite{mocap2003data}. 
However, covariate shift can emerge because of the different statistics of rendered and natural images. 

Since both of these two methods use the same SMPL surface model as the one we use in our work, we can directly compare results, and also combine datasets. We render our dense coordinates and our dense part labels on the SMPL model for all 8514 images of UP dataset and 60k SURREAL models for comparison.

In \reffig{fig:datasets} we assess the test performance of ResNet-101 FCNs of stride 8 trained with different datasets, using a Deeplab-type architecture. During training we augment samples from all of the datasets with scaling, cropping and rotation.
We observe that the surrogate datasets lead to weaker performance, while their  combination yields improved results. Still, their performance is substantially lower than the one obtained by training on our DensePose dataset, while combining the DensePose with SURREAL results in a moderate drop in network performance. 
Based on these results we rely exclusively on the DensePose dataset for training in the remaining experiments, even though domain adaptation could be used in the future \cite{GaninL15} to exploit synthetic sources of supervision.

\newcommand{\wdt}{1}
\begin{figure*}[tp]
\centering
\includegraphics[width=1.\textwidth]{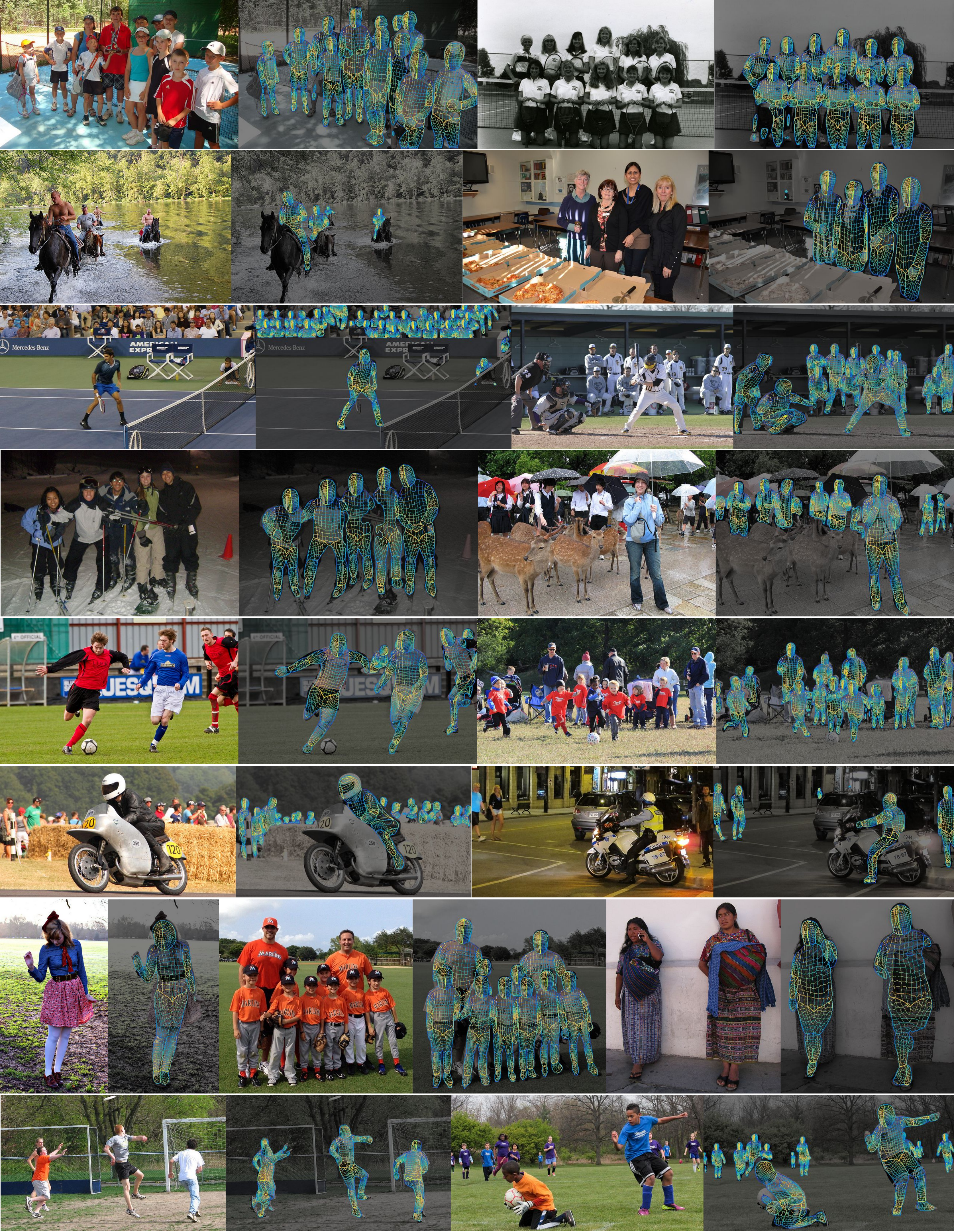}
\caption{ Qualitative evaluation of  DensePose-RCNN. \textit{Left:} input,  \textit{Right:} DensePose-RCNN estimates. We observe that our system successfully estimates body pose regardless of skirts or dresses, while handling 
 a large variability of scales, poses, and occlusions. }
\label{fig:qualitative1}
\end{figure*}

The last line in the table of \reffig{fig:datasets} ('DensePose$^*$') indicates an additional performance boost that we get by using the COCO human segmentation masks in order to  replace background intensities with an average intensity during both training and testing and also by evaluating  the network at multiple scales and averaging the results. Clearly, the results with other methods are not directly comparable, since we are using additional information to remove background structures. Still, the resulting predictions are substantially closer to human performance -- we therefore use this as the `teacher network'  to obtain dense supervision for the experiments in \refsec{second}. 

\subsubsection{FCNN- vs Model-based pose estimation}
\label{modelc}
 In \reffig{fig:SMPLFY}
we compare our method to the SMPLify pipeline of \cite{smpl3d}, which fits the 3D SMPL model to an image based on a pre-computed set of landmark points. We use the code provided by \cite{UP} with both DeeperCut pose estimation landmark detector \cite{deepercut} for 14-landmark results and with the 91-landmark alternative proposed in \cite{UP}. Note that these landmark detectors were trained on the MPII dataset. Since the whole body is visible in the MPII dataset, for a fair comparison we separately evaluate on images where 16/17 or 17/17 landmarks are visible and on the whole test set. 
We observe that while being orders of magnitude faster (0.04-0.25'' vs 60-200'') our bottom-up, feedforward method largely outperforms the iterative, model fitting result. As mentioned above, this difference in accuracy indicates the merit of having at our disposal DensePose-COCO for discriminative training. 


\subsection{Multi-Person Dense Pose Estimation}
\label{secondpart}
\label{second}

Having established the merit of the DensePose-COCO dataset, we now turn to examining the impact of network architecture on dense pose estimation in-the-wild. In
\reffig{fig:dense_errors} we summarize our experimental findings using the same RCP measure used in \reffig{fig:datasets}.

We observe firstly that the FCN-based performance in-the-wild (curve `DensePose-FCN') is now dramatically lower than that of the DensePose curve in \reffig{fig:dense_errors}. Even though we apply a multi-scale testing strategy that fuses probabilities from multiple runs using input images of different scale~\cite{PSP}, the FCN is not sufficiently robust to deal with the variability in object scale.    

We then observe in curve `DensePose-RCNN' a big boost in performance thanks to switching to a region-based system. The networks up to here have been trained using the sparse set of points that have been manually annotated. In curve `DensePose-RCNN-Distillation' we see that using the dense supervision signal delivered by our DensePose$^*$ system on the training set yields a substantial improvement. Finally, in `DensePose-RCNN-Cascade' we show the performance achieved thanks to the introduction of cascading: \refsec{cascade} almost matches the 'DensePose$^*$' curve of \reffig{fig:datasets}. 

\newcommand{\te}[1]{#1}

\begin{table*}[h!]
\begin{center}
\begin{tabular}{|l|p{9mm}p{9mm}p{9mm}|p{9mm}p{9mm}|p{9mm}p{9mm}p{9mm}|p{9mm}p{9mm}|}
\hline
\hfill \emph{Method} \hfill\, &  \emph{$\mathbf{AP}$} & \emph{$\mathbf{AP}_{50}$} & \emph{$\mathbf{AP}_{75}$} & \emph{$\mathbf{AP}_{M}$} & \emph{$\mathbf{AP}_{L}$}  & \emph{$\mathbf{AR}$} & \emph{$\mathbf{AR}_{50}$} & \emph{$\mathbf{AP}_{75}$} & \emph{$\mathbf{AR}_{M}$} & \emph{$\mathbf{AR}_{L}$}\\
\hline\hline
$\vphantom{\rule{0mm}{3.5mm}}$%
DensePose (ResNet-50)& \te{51.0} & \te{83.5} & \te{54.2} & \te{39.4} & \te{53.1} & \te{60.1} & \te{88.5} & \te{64.5} & \te{42.0} & \te{61.3}\\ 
DensePose (ResNet-101) & \te{51.8} & \te{83.7} & \te{56.3} & \te{42.2} & \te{53.8} & \te{61.1} & \te{88.9} & \te{66.4} & \te{45.3} & \te{62.1}\\ 
\hline\hline
\multicolumn{11}{|c|}{$\vphantom{\rule{0mm}{4mm}}$%
\emph{Multi-task learning}}\\ \hline
$\vphantom{\rule{0mm}{3.5mm}}$
DensePose + masks & \te{51.9} & \te{85.5} & \te{54.7} & \te{39.4} & \te{53.9} & \te{61.1} & \te{89.7} & \te{65.5} & \te{42.0} & \te{62.4} \\
DensePose + keypoints & \te{52.8} & \te{85.6} & \te{56.2} & \te{42.2} & \te{54.7} & \te{62.6} & \te{89.8} & \te{67.7} & \te{45.4} & \te{63.7}\\ 
\hline\hline
\multicolumn{11}{|c|}{$\vphantom{\rule{0mm}{4mm}}$%
\emph{Multi-task learning with cascading}}\\ \hline
$\vphantom{\rule{0mm}{3.5mm}}$%
DensePose-cascade & \te{51.6} & \te{83.9} & \te{55.2} & \te{41.9} & \te{53.4} & \te{60.4} & \te{88.9} & \te{65.3} & \te{43.3}  & \te{61.6} \\
DensePose + masks & \te{52.8} & \te{85.5} & \te{56.1} & \te{40.3} & \te{54.6} & \te{62.0} & \te{89.7} & \te{67.0} & \te{42.4} & \te{63.3} \\
DensePose + keypoints & \te{55.8} & \te{87.5} & \te{61.2} & \te{48.4} & \te{57.1} & \te{63.9} & \te{91.0} & \te{69.7} & \te{50.3} & \te{64.8} \\
\hline
\end{tabular}
\caption{  Per-instance evaluation of  DensePose-RCNN performance on COCO \texttt{minival} subset. All multi-task experiments are based on ResNet-50 architecture. DensePose-cascade corresponds to the base architecture with an iterative refinement module with no input from other tasks.}

\label{tab:coco_multitask_dp}

\end{center}
\end{table*}

This is a remarkably positive result: as described in \refsec{firstpart}, the `DensePose$^*$' curve corresponds to a very privileged evaluation, involving (a) cropping objects around their ground-truth boxes and fixing their scale (b) removing background variation from both training and testing, by using ground-truth object masks and (c) ensembling over scales. It can therefore be understood as an upper bound of what we could expect to obtain when operating in-the-wild. We see that our best system is marginally below that level of performance, which clearly reveals the power of the three modifications we introduce, namely region-based processing, inpainting the supervision signal, and cascading.

\medskip
In \reftab{tab:coco_multitask_dp} we report the AP and AR metrics described in \refsec{supervision} as we change different choices in our architecture. 
We have conducted experiments using both ResNet-50 and ResNet-101 backbones and observed an only insignificant boost in performance with the larger model (first two rows in \reftab{tab:coco_multitask_dp}). The rest of our experiments are therefore based on the ResNet-50-FPN version of DensePose-RCNN. 
The following two experiments shown in the
middle section of \reftab{tab:coco_multitask_dp} indicate the impact 
on multi-task learning. 

Augmenting the network with the mask or keypoint branches yields improvements with any of these two auxiliary tasks.
%
The last section of \reftab{tab:coco_multitask_dp} reports improvements in dense pose estimation obtained through  cascading using the network setup from \reffig{fig:Cascade}.
Incorporating additional guidance in particular from the keypoint branch significantly boosts performance.

\subsection{Qualitative Results}
\label{secondpart}

In this section we provide additional qualitative results to further demonstrate the performance of our method. In \reffig{fig:qualitative1} we show qualitative results generated by our method, where the correspondence is visualized in terms of `fishnets', namely isocontours of estimated UV coordinates that are superimposed on humans. 
As these results indicate, our method is able to handle large amounts of occlusion, scale, and pose variation, while also successfully hallucinating the human body behind clothes such as dresses or skirts. 

\begin{figure*}[!h]
\centering
\includegraphics[width=0.9\textwidth]{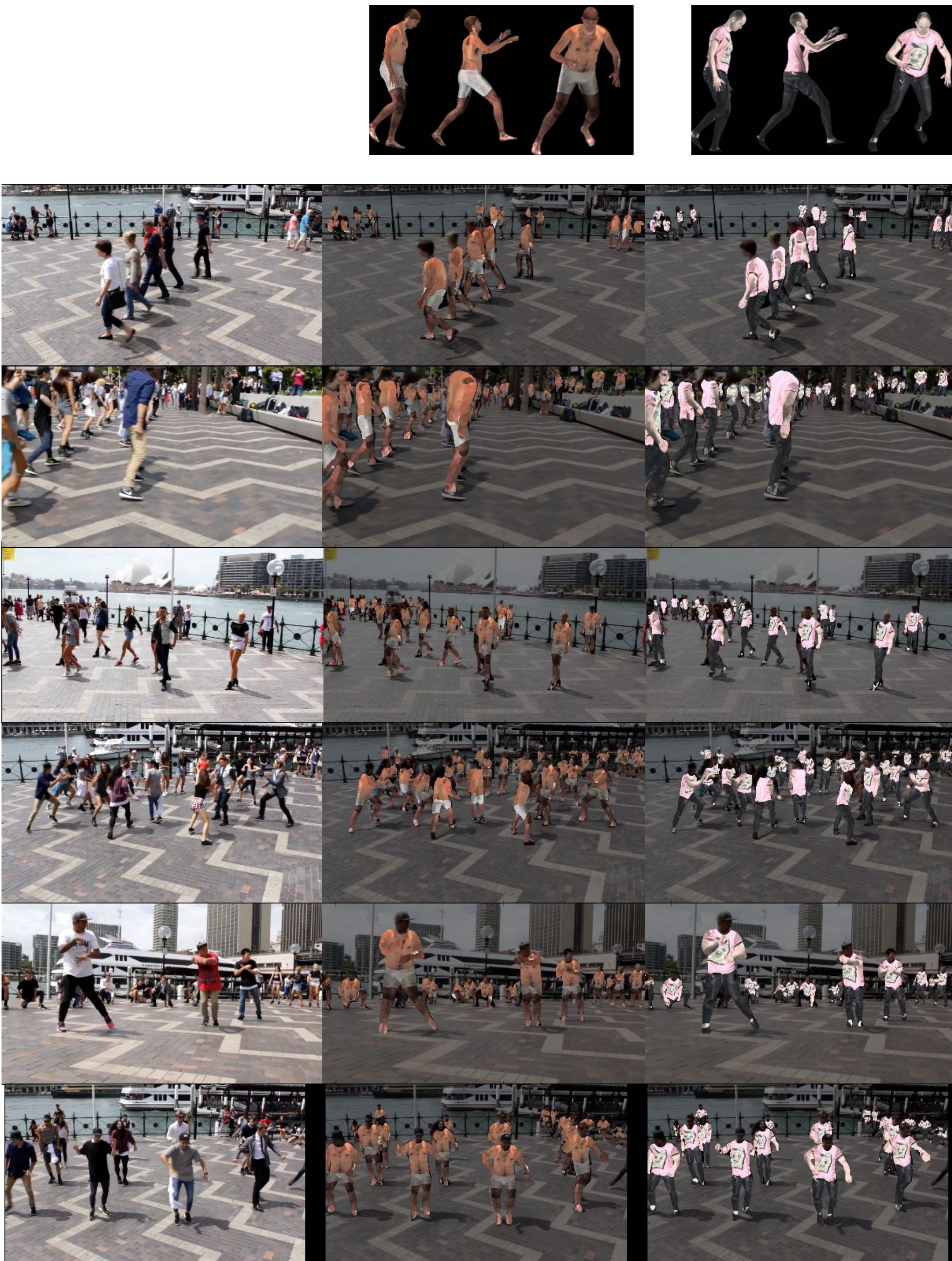}
\caption{ Qualitative results for texture transfer: The textures that are provided in the top row are mapped to image pixels based on estimated correspondences. The whole video can be seen at \url{http://densepose.org}.}
\label{fig:transfer}
\end{figure*}

In Fig.\ref{fig:transfer} we demonstrate a simple graphics-oriented application, where we map  texture RGB intensities taken from \cite{varol2017learning} to estimated UV body coordinates - the whole video is available on our project's website \url{http://densepose.org}.

\section{Conclusion}
In this work we have tackled the task of dense human pose estimation using discriminative trained models. We have introduced COCO-DensePose, a large-scale dataset of ground-truth image-surface correspondences
and developed novel architectures that allow us to recover highly-accurate dense correspondences between images and the body surface in multiple frames per second. We anticipate that this will pave the way both for downstream tasks in  augmented reality or graphics, but also help us tackle the general problem of associating images with  semantic 3D object representations. 

\section*{Acknowledgements}
We thank the authors of \cite{maskRCNN} for sharing their code, Piotr Dollar for guidance and proposals related to our dataset's quality,  Tsung-Yi Lin for his help with COCO-related issues and H. Yi\u{g}it G\"uler for his help with backend development. 

{\small
\bibliographystyle{ieee}
\bibliography{egbib}
}

\end{document}